\documentclass[10pt,twocolumn,letterpaper]{article}

\usepackage{iccv}              

\definecolor{iccvblue}{rgb}{0.21,0.49,0.74}
\usepackage{enumitem}
\usepackage{multirow}
\usepackage{appendix}
\usepackage[T1]{fontenc}
\usepackage{enumitem}
\usepackage{amsmath}
\usepackage{amssymb}
\usepackage{multirow}
\usepackage{tabularx}
\usepackage{booktabs}
\usepackage{makecell}
\usepackage{arydshln}
\usepackage{pifont}

\usepackage[pagebackref,breaklinks,colorlinks,allcolors=iccvblue]{hyperref}

\def\confName{ICCV}
\def\confYear{2025}

\title{SpiLiFormer: Enhancing Spiking Transformers with Lateral Inhibition}

\author{
Zeqi Zheng\textsuperscript{1,2*}, 
Yanchen Huang\textsuperscript{2,3*},
Yingchao Yu\textsuperscript{4,2},
Zizheng Zhu\textsuperscript{2,5}, \\
Junfeng Tang\textsuperscript{1,2},
Zhaofei Yu\textsuperscript{6},
Yaochu Jin\textsuperscript{2}$^\dagger$ \\
\textsuperscript{1}Zhejiang University \quad 
\textsuperscript{2}Westlake University \quad 
\textsuperscript{3}Nanjing University \quad 
\textsuperscript{4}Donghua University\\
\textsuperscript{5}University of Electronic Science and Technology of China \quad
\textsuperscript{6}Peking University\\
\tt\small \{zhengzeqi, huangyanchen, jinyaochu\}@westlake.edu.cn}

\begin{document}
\maketitle
{\renewcommand{\thefootnote}{}
\footnotetext{$^*$ Equal contribution. $^\dagger$ Corresponding author.}}
\begin{abstract}
Spiking Neural Networks (SNNs) based on Transformers have garnered significant attention due to their superior performance and high energy efficiency. However, the spiking attention modules of most existing Transformer-based SNNs are adapted from those of analog Transformers, failing to fully address the issue of over-allocating attention to irrelevant contexts. To fix this fundamental yet overlooked issue, we propose a Lateral Inhibition-inspired Spiking Transformer (SpiLiFormer). It emulates the brain's lateral inhibition mechanism, guiding the model to enhance attention to relevant tokens while suppressing attention to irrelevant ones. Our model achieves state-of-the-art (SOTA) performance across multiple datasets, including CIFAR-10 (+0.45\%), CIFAR-100 (+0.48\%), CIFAR10-DVS (+2.70\%), N-Caltech101 (+1.94\%), and ImageNet-1K (+1.6\%)~\footnote{These results are obtained by comparing with state-of-the-art spiking neural network models that have a similar number of parameters.}. Notably, on the ImageNet-1K dataset, SpiLiFormer (69.9M parameters, 4 time steps, 384 resolution) outperforms E-SpikeFormer (173.0M parameters, 8 time steps, 384 resolution), a SOTA spiking Transformer, by 0.46\% using only 39\% of the parameters and half the time steps. The code and model checkpoints are publicly available at https://github.com/KirinZheng/SpiLiFormer. 

\end{abstract}    
\section{Introduction}
\label{sec:intro}

Spiking Neural Networks (SNNs), regarded as the third generation of neural networks~\cite{maass1997networks}, are seen as a potential alternative to Artificial Neural Networks (ANNs) due to their biological interpretability and high energy efficiency, which stem from their event-driven properties. 
Transformer~\cite{vaswani2017attention}, originally designed for natural language processing tasks, has now become a dominant neural network architecture, demonstrating remarkable performance across various visual tasks~\cite{dosovitskiy2020image, yuan2021tokens, zhu2020deformable, wang2021pyramid}.
The success of Transformers has driven the exploration of their integration with SNNs, with the objective of enhancing SNN models' performance across various tasks and progressively bridging the performance gap with ANNs, particularly in image classification~\cite{zhou2022spikformer, yao2024spikedriven, yao2024spikedrivenv2, zhou2024qkformer}, object detection~\cite{luo2024integer, kim2020spiking}, action recognition~\cite{liu2021event, vicente2025spiking}, and semantic segmentation~\cite{long2024spike, yao2024spikedrivenv2}.

The attention module plays a crucial role in Transformer-based SNNs, significantly influencing model performance.
Recent research advancements in spiking attention modules can be broadly classified into three categories: 
1) Self-Spiking Attention (SSA)~\cite{zhou2022spikformer, zhou2024spikformer, zhou2023spikingformer}, which directly converts the Query ($Q$), Key ($K$), and Value ($V$) into sparse spikes; 
2) Spike-Driven Self-Attention (SDSA)~\cite{yao2024spikedriven, yao2024spikedrivenv2} and Spiking RWKV (S-RWKV)~\cite{zhu2023spikegpt}, which replace the dot product in SSA with the Hadamard product, achieving linear complexity while reducing both energy consumption and computational cost;
3) Spatial-Temporal Attention (STA)~\cite{wang2023spatial, lee2024spiking}, which refines traditional spatial attention by explicitly modeling the key temporal dependencies intrinsic to spike-based processing.

\begin{figure*}[t]
    \centering
    \includegraphics[width=0.99\textwidth]{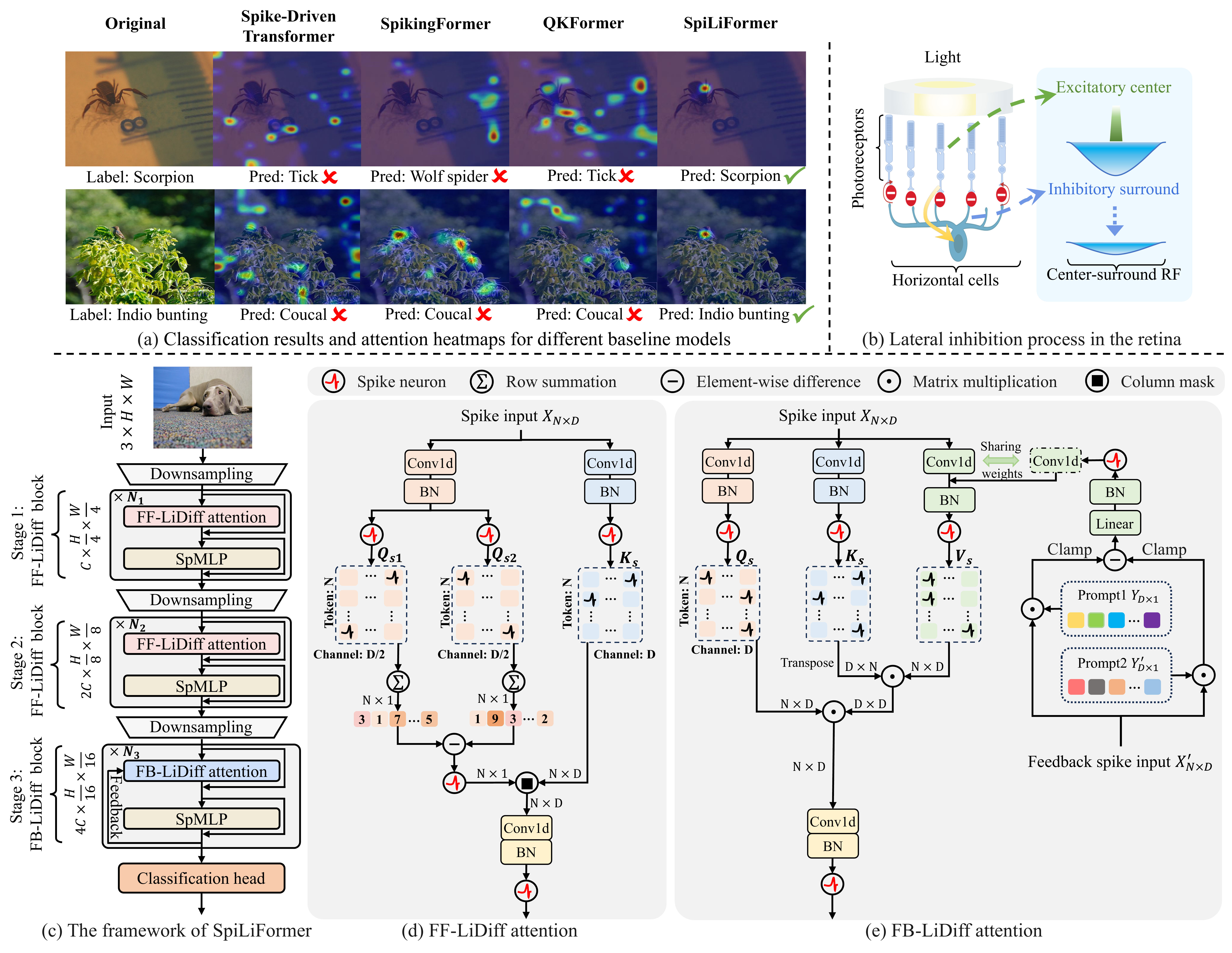}
    \caption{
    Illustration of the severe attention distraction phenomenon in Transformer-based SNNs, the architecture of the Lateral Inhibition-inspired Spiking Transformer (SpiLiFormer), and the detailed processing mechanism of Feedfoward-pathway Lateral Differential Inhibition (FF-LiDiff) attention and Feedback-pathway Lateral Difference Inhibition (FB-LiDiff) attention.
    (a) refers to the attention distraction in mainstream Transformer-based SNNs, where models excessively focus on irrelevant background information, leading to misclassification.
    (b) denotes the retinal lateral inhibition mechanism, illustrating how horizontal cells regulate neural responses to enhance contrast and suppress noise.
    (c) presents the architecture of SpiLiFormer. (d) and (e) represent the information processing flows of FF-LiDiff and FB-LiDiff attention, respectively.}
    \label{architecture}
\end{figure*}

However, we observe that most Transformer-based SNNs exhibit a phenomenon referred to as \textbf{attention distraction}. 
As illustrated in \cref{architecture}(a), the model disproportionately assigns high attention weights to background information during decision-making, leading to the neglect of critical object-related features and ultimately impairing classification performance. 
This issue stems from the fact that spiking attention largely inherits the traditional attention forward propagation paradigm. 
Previous studies~\cite{chefer2022optimizing, zhang2025selective, sahiner2022unraveling} suggest that traditional attention mechanisms process $Q$, $K$, and $V$ uniformly in the mapping $softmax(QK^\top)V$, thereby limiting their capacity to regulate contextual sparsity and relevance.
Therefore, we posit that constructing additional pathways to process $Q$, $K$, and $V$ separately is essential for mitigating attention distraction in Transformer-based SNNs and enhancing model performance.

By contrast, the natural lateral inhibition mechanism in the visual system enables the brain to focus on important areas, thereby improving perception and reducing visual overload~\cite{shaw1975retinal, magnuson2024contra, del2024lateral}. Specifically, when neurons within a given area are strongly activated, adjacent neurons are inhibited. This principle is first observed in the retina, where horizontal cells receive input from photoreceptors and inhibit both the stimulated photoreceptors and their neighbors, forming a center-surround receptive field (RF), as shown in \cref{architecture}(b). In higher visual areas, such as the primary (V1) and secondary (V2) visual cortices, a similar mechanism operates through long-range connections, which contrasts with the short-range inhibition in the retina, helping to refine visual processing by suppressing irrelevant stimuli and selectively enhancing salient features~\cite{zhu2024hierarchies, znamenskiy2024functional}.

Based on the short-range and long-range lateral inhibition mechanisms discussed above, we propose the Lateral Inhibition-inspired Spiking Transformer (SpiLiFormer) to address the attention distraction issue in Transformer-based SNNs, as shown in \cref{architecture}(c).
This model consists of two types of stacked modules: 1) Feedforward-pathway Lateral Differential Inhibition (FF-LiDiff) blocks and 2) Feedback-pathway Lateral Differential Inhibition (FB-LiDiff) blocks. Moreover, it introduces the following innovations: 
1) In the shallow blocks (\ie, Stage 1 and Stage 2), it explicitly incorporates an \textbf{inhibition differential attention mechanism} by emulating the retinal short-range lateral inhibition process (see \cref{architecture}(d) for details);
2) In the deep blocks (\ie, Stage 3), it employs \textbf{feedback-driven processing based on differential attention} to simulate the long-range lateral inhibition mechanism observed in the cerebral cortex (see \cref{architecture}(e) for details).
We evaluate our proposed SpiLiFormer on five image classification datasets, including CIFAR-10~\cite{krizhevsky2009learning}, CIFAR-100~\cite{krizhevsky2009learning}, CIFAR10-DVS~\cite{li2017cifar10}, N-Caltech101~\cite{orchard2015converting}, and ImageNet-1K~\cite{deng2009imagenet}, demonstrating superior performance over current SOTA models. Additionally, adversarial testing and attention heatmap visualizations on the ImageNet-1K test set and noise-augmented datasets (\ie, CIFAR-10-C and ImageNet-1K-C~\cite{hendrycks2019benchmarking}) show that SpiLiFormer effectively suppresses attention on irrelevant areas and background noise. The main contributions of this work are:

\begin{itemize}[label=\textbullet] 
    \item We propose SpiLiFormer, which incorporates a brain-inspired lateral inhibition mechanism and introduces two novel attention paradigms: \textbf{FF-LiDiff attention} and \textbf{FB-LiDiff attention}. This model effectively mitigates the attention distraction issue common in current Transformer-based SNNs, thereby enhancing image classification performance.
    \item We evaluate SpiLiFormer across multiple datasets, achieving new SOTA performance on CIFAR-10, CIFAR-100, CIFAR10-DVS, N-Caltech101, and ImageNet-1K. Specifically, on ImageNet-1K, SpiLiFormer (69.9M parameters, 4 time steps) outperforms the optimal SOTA model, E-SpikeFormer (173.0M parameters, 8 time steps), by 0.46\% in top-1 accuracy, while using only 0.39 times the number of parameters and half the time steps.
    \item We comprehensively evaluate the robustness of SpiLiFormer through attention heatmap visualizations and adversarial testing, and find that SpiLiFormer achieves more effective attention allocation and enhanced robustness compared to other baseline spiking models.
\end{itemize}

\section{Related Work}

\subsection{Transformer-based SNNs}

Although convolution-based SNNs exhibit high energy efficiency, they still demonstrate a significant performance gap compared to ANNs. 
To address this challenge, researchers have explored integrating Transformer architectures with SNNs. 
Spikformer~\cite{zhou2022spikformer} was the first to incorporate the Transformer attention into SNNs, cleverly replacing the Softmax operation by leveraging the binary characteristic of spike representations to encode Query, Key, and Value. 
Building upon this, Spikingformer~\cite{zhou2023spikingformer} introduced a pre-activation shortcut to avoid floating-point multiplication while simultaneously reducing the spike firing rate. 
Inspired by SEW-ResNet~\cite{fang2021deep}, Spike-driven Transformer~\cite{yao2024spikedriven} replaced the dot product with a Hadamard product and reshaped the spike residual connection based on the membrane potential to further minimize energy consumption and leverage the event-driven property. 
Additionally, Spiking RWKV~\cite{zhu2023spikegpt} and QKFormer~\cite{zhou2024qkformer} introduced linear attention into SNNs to reduce computational complexity. 
Meanwhile, STA-Transformer~\cite{lee2024spiking} effectively integrated both spatial and temporal dynamics of spike trains to enhance model performance.
However, most existing Transformer-based SNNs still adhere to the traditional attention propagation paradigm of ANNs, which results in the severe phenomenon of attention distraction, thereby limiting model performance.

\subsection{Models with Lateral Inhibition}

Lateral inhibition mechanisms have been widely adopted in various SNN applications. In speech recognition, Spiking-LEAF~\cite{song2024spiking} emulated inner hair cell functionality and leveraged lateral inhibition feedback to enhance the neuron model, significantly improving the efficiency and noise robustness of spike encoding.
In the object recognition, LISNN~\cite{cheng2020lisnn} modeled lateral inhibition by establishing spatial neighbor-order relationships between neurons, thereby enhancing the model's ability to focus on important features. 
In the image classification, Zhang \etal~\cite{zhang2025biologically} proposed an adaptive self-organizing lateral inhibition strategy, where inhibition strength was adjusted according to the Euclidean distance between neurons. 
This approach addressed the issue of inefficient feature clustering and reduced redundancy in neuron activation. 
However, existing research on lateral inhibition has predominantly focused on small-scale datasets for validating model feasibility, with limited exploration of its applicability to larger and more complex datasets. Furthermore, its integration with Transformer-based SNNs remains largely unexplored.

\section{Method}

This section details the proposed SpiLiFormer model, covering the spiking neuron layer, overall architecture, and two lateral inhibition-inspired attention mechanisms. Additionally, we outline the training strategy and loss function.

\subsection{Spiking Neuron Layer}

The spiking neuron layer relies on cumulative firing to integrate spatio-temporal information into the membrane potential, which is then converted into binary spikes to drive computation in the next layer.
We adopt the widely used Leaky Integrate-and-Fire (LIF) spiking neuron layer, as established in previous studies~\cite{yao2024spikedriven, zhou2024qkformer, yao2025scaling}, with the following dynamics:

\begin{equation}
  U[t] = H[t](1-S[t]) + U_{reset}S[t],
  \label{eq:LIF_dynamic_1}
\end{equation}

\begin{equation}
  H[t]=U[t-1] + \frac{1}{\tau}(X[t]-(U[t-1]-U_{reset})),
  \label{eq:LIF_dynamic_2}
\end{equation}

\begin{equation}
  S[t] = \Theta(H[t]-U_{th}),
  \label{eq:LIF_dynamic_3}
\end{equation}
where $U_{\text{reset}}$ represents the reset potential, and $X[t]$ and $U[t]$ denote the current input and membrane potential at timestep $t$, respectively. $\Theta(\cdot)$ is the Heaviside step function, which outputs 1 if $x \geq 0$ and 0 otherwise. When $H[t]$ exceeds the firing threshold $U_{th}$, the spiking neuron fires a spike $S[t]$, and $U[t]$ is reset to $U_{reset}$ at that moment. Additionally, $\tau$ represents the membrane time constant, which controls the rate of membrane potential leakage.

\subsection{Overall Architecture}

The overall architecture of SpiLiFormer is illustrated in~\cref{architecture}(c), which consists of three stages: Stage 1, Stage 2, and Stage 3, forming a three-level hierarchical structure. 
The input data is represented as $I \in \mathbb{R}^{T \times N \times H \times W}$, where $N=3$ for static image data and $N=2$ for neuromorphic data. 
In Stage 1, a $ 4 \times 4 $ patch size is utilized, mapping each patch's input feature dimension ($4 \times 4 \times N$) to a spike-form representation of arbitrary dimension ($C$) via the Downsampling module. 
This process reduces the number of tokens to $\frac{H}{4} \times \frac{W}{4}$. The transformed features are then processed through FF-LiDiff attention and a Spiking Multi-Layer Perceptron (SpMLP). FF-LiDiff attention introduces an additional forward pathway and employs a differential approach to mitigate attention distraction in the network's shallower layers (see \cref{sec:FF-LiDiff Attention} for details).
In Stage 2, the processing mechanism is similar to Stage 1, but with a $2 \times 2$ patch size, an expanded channel dimension of $2C$, and the number of tokens reduced to $\frac{H}{8} \times \frac{W}{8}$. 
Finally, in Stage 3, we retain the $2 \times 2$ patch size, expand the channel dimension to $4C$, and reduce the number of tokens to $\frac{H}{16} \times \frac{W}{16}$. Additionally, a more complex FB-LiDiff attention module with feedback loops is introduced (see \cref{sec:FB-LiDiff Attention} for details) to replace the previous FF-LiDiff attention, further mitigating attention distraction in the model. 

The number of spiking blocks (either FF-LiDiff block or FB-LiDiff block) in each stage is denoted as $N_1$, $N_2$, and $N_3$, respectively. Since the Downsampling, SpMLP and classification head modules are based on prior research works~\cite{yao2024spikedrivenv2, zhou2024qkformer, yao2025scaling}, the following sections will focus on the detailed description of FF-LiDiff attention and FB-LiDiff attention.

\subsection{Feedforward-pathway Lateral Differential Inhibition Attention}
\label{sec:FF-LiDiff Attention}

An overview of FF-LiDiff attention is shown in Figure \ref{architecture}(d). For better mathematical description, in subsequent discussion, we assume $T=1$ and only use single head attention. Given a spike input $X \in \mathbb{R}^{N \times D}$ ($N$ is the token number, $D$ is the channel number), Query ($Q$) and Key ($K$) can be computed through learnable matrices:
\begin{equation}
  Q= BN(XW_Q), K = BN(XW_K),
  \label{eq:FF_LiDiff_attn_1}
\end{equation}
where $W_Q, W_K \in \mathbb{R}^{D \times D}$ and $Q, K \in \mathbb{R}^{N \times D}$. 

To implement the lateral inhibition mechanism, we first split $ Q $ along the channel dimension into two parts, obtaining $ Q_1 $ and $ Q_2 $, before computing the attention matrix. Then, all of $Q_1$, $Q_2$, and $K$ are transformed into spike trains via their respective spiking neuron layers:
\begin{equation}
  Q_{s1}=\mathcal{SN}(Q_1), Q_{s2}=\mathcal{SN}(Q_2), K_{s}= \mathcal{SN}(K),
  \label{eq:FF_LiDiff_attn_2}
\end{equation}
where $Q_{s1}, Q_{s2} \in \mathbb{R}^{N \times \frac{D}{2}}$, $ K_s \in \mathbb{R}^{N \times D}$. Next, we sum along the channel dimension to obtain the excitatory attention $A_e$ and inhibitory attention $A_i$. Specifically, $A_e$ and $A_i$ are $N \times 1$ token attention vectors, representing the importance of different tokens from the perspectives of excitation and inhibition, respectively. Finally, we compute the element-wise difference between $A_e$ and $A_i$, and pass the aggregated attention through a spiking neural layer to obtain binary attention $A_{\text{combined}}$. 
This attention is then applied as a token-wise mask to $K_s$ via the Hadamard product $\otimes$. The above process can be described as:
\begin{equation}
  A_e= \sum_{j}^{\frac{D}{2}} Q_{s1}^{i,j},   A_i = \sum_{j}^{\frac{D}{2}} Q_{s2}^{i,j},
  \label{eq:FF_LiDiff_attn_3}
\end{equation}

\begin{equation}
    A_{\text{combined}} = \mathcal{SN}(A_e-A_i),
    \label{eq:FF_LiDiff_attn_4}
\end{equation}

\begin{equation}
    X' = A_{\text{combined}} \otimes K_s.
    \label{eq:FF_LiDiff_attn_5}
\end{equation}

\subsection{Feedback-pathway Lateral Differential Inhibition Attention}
\label{sec:FB-LiDiff Attention}

FB-LiDiff attention requires two forward propagations as shown in Figure \ref{architecture}(e). Specifically, during the first forward propagation, FB-LiDiff attention takes spike-form $S_{\text{stage3}}^{Ds}$ from the Downsampling module in Stage 3 or takes spike-form $S_{\text{stage3}}^{i}$ ($i \in [1, N_3]$) from the $i$-th preceding FB-LiDiff block as input. Given an input $X_1^{m}$ to the $m$-th FB-LiDiff attention, the first forward propagation can be described as follows:
\begin{equation}
    Y_s^{1,m} = \mathcal{SN}(BN(X_1^{m} W_Y^{m})), Y \in \{Q, K, V\},
    \label{eq:FB_LiDiff_attn_1}
\end{equation}

\begin{equation}
    Attn^{1,m} = Q_s^{1,m} \odot ((K_s^{1,m})^\top \odot V_s^{1,m}),
    \label{eq:FB_LiDiff_attn_2}
\end{equation}

\begin{equation}
   O_{\text{stage3}}^{m} = \mathcal{SN}(BN(Attn^{1,m} W_{attn}^{m})),
    \label{eq:FB_LiDiff_attn_3}
\end{equation}
where $X_1^{m} \in \mathbb{R}^{N \times D}$ ($X_1^{m} \in \{ S_{\text{stage3}}^{Ds}, S_{\text{stage3}}^{i} \}$), and $Y_s^{1,m}$ ($Y \in \{Q, K, V\}$) represents the corresponding spiking representation of $Q$, $K$ and $V$ in the $m$-th FB-LiDiff attention. $W_z^{m}$ ($z \in \{Q, K, V, attn\}$) and $O_{\text{stage3}}^{m}$ denote the learnable weight matrix and the spike-form output of the $m$-th FB-LiDiff attention, respectively. $\odot$ refers to the dot product.

Before the second forward propagation, we use the output $ O_{\text{stage3}}^{N_3} $ from the last block of Stage 3 as feedback information and return it to each FB-LiDiff attention. To filter the feedback information and further mitigate the attention distraction problem, we first introduce learnable prompts $ W_{p1} $ and $ W_{p2} $, where $ W_{p1}, W_{p2} \in \mathbb{R}^{1 \times D} $. Next, we compute the dot product between $ W_{p1} $ and $ O_{\text{stage3}}^{N_3} $, and similarly between $ W_{p2} $ and $ O_{\text{stage3}}^{N_3} $. These dot products are then broadcast along the token dimension. The results are scaled to the range $[0,1]$ using the $ \text{clamp}(\cdot) $ function, forming the excitatory feedback attention $ A_e^{\text{FB}} $ and inhibitory feedback attention $ A_i^{\text{FB}} $. Subsequently, we compute the element-wise difference between $ A_e^{\text{FB}} $ and $ A_i^{\text{FB}} $ to obtain the aggregated attention $ A_{\text{combined}}^{\text{FB}} $. Finally, we apply a linear transformation to $ A_{\text{combined}}^{\text{FB}} $ and pass it through the spiking neuron layer to align it with the feature space of different FB-LiDiff attentions. These processes are formulated as:

\begin{equation}
    A_e^{\text{FB}} = \text{clamp}(O_{\text{stage3}}^{N_3} \odot W_{p1}, 0, 1),
    \label{eq:FB_LiDiff_attn_4}
\end{equation}

\begin{equation}
    A_i^{\text{FB}} = \text{clamp}(O_{\text{stage3}}^{N_3} \odot W_{p2}, 0, 1),
    \label{eq:FB_LiDiff_attn_5}
\end{equation}

\begin{equation}
    A_{\text{combined}}^{\text{FB}} = A_e^{\text{FB}} - A_i^{\text{FB}},
    \label{eq:FB_LiDiff_attn_6}
\end{equation}

\begin{equation}
    X_{\text{FB}}^{m} = \mathcal{SN}(BN(A_{\text{combined}}^{\text{FB}} W_{\text{FB}}^{m})).
    \label{eq:FB_LiDiff_attn_7}
\end{equation}

The second forward propagation begins by taking the processed forward representation \( S_{\text{stage3}}^{Ds} \) and the corresponding feedback information \( X_{\text{FB}}^{1} \) as input to the first FB-LiDiff attention. The objective is to reduce model energy consumption by eliminating redundant computations in the non-feedback loop (\ie., Stage 1 and Stage 2). A detailed energy consumption analysis can be found in Appendix~A.3. Additionally, during the second forward propagation, we share all parameter weights across the FB-LiDiff blocks to minimize the number of model parameters. Given the input $ X_2^{m} $, the process described above can be formulated as:

\begin{equation}
    Y_s^{2,m} = \mathcal{SN}(BN(X_2^{m} W_Y^{m})), Y \in \{Q, K\},
    \label{eq:FB_LiDiff_attn_8}
\end{equation}

\begin{equation}
    V_s^{2,m} = \mathcal{SN}(BN(X_2^{m} W_V^{m} + X_{\text{FB}}^{m}W_V^{m})),
    \label{eq:FB_LiDiff_attn_9}
\end{equation}

\begin{equation}
    Attn^{2,m} = Q_s^{2,m} \odot ((K_s^{2,m})^\top \odot V_s^{2,m}), 
    \label{eq:FB_LiDiff_attn_10}
\end{equation}

\begin{equation}
    (O_{\text{stage3}}^{m})^{'} = \mathcal{SN}(BN(Attn^{2,m} W_{attn}^{m})).
    \label{eq:FB_LiDiff_attn_11}
\end{equation}

\subsection{Training Strategy and Loss Function}

We employ a surrogate gradient-based method to overcome the non-differentiability of the spiking neuron activation in \cref{eq:LIF_dynamic_3}, enabling direct training of the model from scratch, as demonstrate in previous works~\cite{zhou2022spikformer, zhou2023spikingformer, zhu2023spikegpt, zhou2024qkformer, yao2024spikedrivenv2}. In order to make better use of the spiking representations obtained from two forward propagations, we adjust the model's loss function as follows:

\begin{equation}
    \mathcal{L}_{1} = \mathcal{L}_{CE}(CH(O_{\text{stage3}}^{N_3}), y),
    \label{eq:loss_function_1}
\end{equation}

\begin{equation}
    \mathcal{L}_{2} = \mathcal{L}_{CE}(CH((O_{\text{stage3}}^{N_3})^{'}), y),
    \label{eq:loss_function_2}
\end{equation}

\begin{equation}
    \mathcal{L}_{\text{SpiLiFormer}} = \alpha \mathcal{L}_{1} + (1-\alpha) \mathcal{L}_{2},
    \label{eq:loss_function_3}
\end{equation}
where $\alpha$ is a hyperparameter that dynamically balances the losses from two forward propagations. Unless explicitly specified otherwise, $\alpha$ is set to 0.5 in the subsequent experiments. The choice of this value is validated through an ablation study, as detailed in Appendix~A.4. Additionally, $y$ represents the true label; $O_{\text{stage3}}^{N_3}$ and $(O_{\text{stage3}}^{N_3})^{'}$ denote the outputs from the first and second forward propagations of the last FB-LiDiff block, respectively; $CH(\cdot)$ represents the classification head module, as described in previous studies~\cite{yao2024spikedriven, yao2024spikedrivenv2}; and $\mathcal{L}_{CE}$ denotes the cross-entropy loss function.

\begin{table*}[t!]
\resizebox{0.99\textwidth}{!}{%
\begin{tabular}{@{}cccccccc@{}}
\toprule
Methods                         & Type & Architecture        & Input Size & Param(M) & Power(mJ) & Time Step & Top-1 Acc(\%) \\ \midrule
ViT~\cite{dosovitskiy2020image}    & ANN  & ViT-B/16            & 384        & 86.59    & 254.84    & 1         & 77.90         \\ \cmidrule(l){2-8} 
\multirow{2}{*}{DeiT~\cite{touvron2021training}} & ANN  & DeiT-B              & 224        & 86.59    & 80.50     & 1         & 81.80         \\ 
                                & ANN  & DeiT-B              & 384       & 86.59    & 254.84    & 1         & 83.10         \\ \cmidrule(l){2-8} 
\multirow{2}{*}{Swin~\cite{liu2021swin}} & ANN  & Swin Transformer-B  & 224        & 87.77    & 70.84     & 1         & 83.50         \\ 
                                & ANN  & Swin Transformer-B  & 384        & 87.77    & 216.20    & 1         & 84.50         \\ \midrule
SEW ResNet~\cite{fang2021deep}        & SNN  & SEW-ResNet-152      & 224        & 60.19    & 12.89     & 4         & 69.26         \\ 
Spikformer~\cite{zhou2022spikformer}   & SNN  & Spikformer-8-768    & 224        & 66.34    & 21.48     & 4         & 74.81         \\  
Spikingformer~\cite{zhou2023spikingformer}  & SNN  & Spikingformer-8-768 & 224        & 66.34    & 13.68     & 4         & 75.85         \\ \midrule
\multirow{3}{*}{S-Transformer~\cite{yao2024spikedriven}}  & SNN  & S-Transformer-8-512 & 224        & 29.68    & 1.13      & 1         & 71.68         \\ 
                                & SNN  & S-Transformer-8-512 & 224        & 29.68    & 4.50      & 4         & 74.57         \\  
                                & SNN  & S-Transformer-8-768 & 288        & 66.34    & 6.09      & 4         & 77.07         \\ \midrule
\multirow{3}{*}{QKFormer~\cite{zhou2024qkformer}}       & SNN  & HST-10-768          & 224        & 64.96    & 38.91     & 4         & 84.22         \\  
                                & SNN  & HST-10-768$^{*}$    & 288        & 64.96    & 64.27     & 4         & 85.25         \\  
                                & SNN  & HST-10-768          & 384        & 64.96    & 113.64    & 4         & 85.65         \\ \midrule
\multirow{3}{*}{E-SpikeFormer~\cite{yao2025scaling}}  & SNN  & E-SpikeFormer       & 224        & 173.0    & 35.6      & 4$^{\star}$ & 84.7          \\ 
                                & SNN  & E-SpikeFormer       & 224        & 173.0    & 54.7      & 8$^{\star}$ & 85.1          \\ 
                                & SNN  & E-SpikeFormer       & 384        & 173.0    & -         & 8$^{\star}$ & 86.2          \\ \midrule
\multirow{4}{*}{\textbf{\makecell{SpiLiFormer \\ (Ours)}}} 
& SNN  & \multicolumn{1}{l}{SpiLiFormer-10-768}     & 224  & 69.10  & 11.77  & 1   & 81.54         \\ 
& SNN  & \multicolumn{1}{l}{SpiLiFormer-10-768}     & 224  & 69.10  & 44.17  & 4   & 85.82         \\ 
& SNN  & \multicolumn{1}{l}{SpiLiFormer-10-768$^{*}$}     & 288  & 69.10  & 73.52  & 4   & 86.62         \\ 
& SNN  & \multicolumn{1}{l}{SpiLiFormer-10-768$^{**}$}     & 384  & 69.10  & 129.45 & 4   & \textbf{86.66} \\ \bottomrule
\end{tabular}
}
\caption{
Performance comparison on ImageNet-1K. "SpiLiFormer-$L$-$D$" represents Lateral Inhibition-inspired Transformer with $L$ blocks and a $D$-dimensional channel. $^{*}$ and $^{**}$ denote input resolutions of $288^2$ and $384^2$ during inference, respectively. $^{\star}$ denotes an integer-valued quantized training approach~\cite{luo2024integer}, in which activation values are restricted to integers during training and later expanded into spike trains during inference.}
\label{tab:imagenet_result}
\end{table*}

\section{Experiments}

We first evaluate the classification performance of SpiLiFormer model on the large-scale ImageNet-1K dataset~\cite{deng2009imagenet}. Next, we assess the performance of SpiLiFormer on smaller-scale static datasets, including CIFAR-10~\cite{krizhevsky2009learning} and CIFAR-100~\cite{krizhevsky2009learning}. Additionally, we evaluate SpiLiFormer on two popular neuromorphic datasets: CIFAR10-DVS~\cite{li2017cifar10} and N-Caltech101~\cite{orchard2015converting}. Finally, we visualize attention heatmaps and conduct adversarial tests to demonstrate SpiLiFormer’s effectiveness in mitigating attention distraction and enhancing robustness. 

\begin{table*}[t]
\renewcommand{\arraystretch}{0.9}
\centering
\resizebox{0.95\textwidth}{!}{
\begin{tabular}{cccccc}
\toprule
Datasets                      & Methods        & Architecture        & Param(M) & Time Step & Top-1 Acc(\%) \\ \midrule
\multirow{5}{*}{CIFAR-10}     & Spikformer~\cite{zhou2022spikformer} & Spikformer-4-384    & 9.32     & 4         & 95.51         \\ 
                              & Spikingformer~\cite{zhou2023spikingformer} & Spikingformer-4-384 & 9.32     & 4         & 95.81         \\ 
                              & S-Transformer~\cite{yao2024spikedriven} & S-Transformer-2-512 & 10.28    & 4         & 95.60         \\ 
                              & QKFormer~\cite{zhou2024qkformer}  & HST-4-384           & 6.74     & 4         & 96.18         \\ 
                              & \textbf{SpiLiFormer (Ours)} & SpiLiFormer-4-384      & 7.04        & 4         & \textbf{96.63}         \\ \midrule
\multirow{5}{*}{CIFAR-100}    & Spikformer~\cite{zhou2022spikformer} & Spikformer-4-384    & 9.32     & 4         & 78.21         \\ 
                              & Spikingformer~\cite{zhou2023spikingformer} & Spikingformer-4-384 & 9.32     & 4         & 79.21         \\ 
                              & S-Transformer~\cite{yao2024spikedriven} & S-Transformer-2-512 & 10.28    & 4         & 78.4          \\ 
                              & QKFormer~\cite{zhou2024qkformer}   & HST-4-384           & 6.74     & 4         & 81.15         \\ 
                              & \textbf{SpiLiFormer (Ours)} & SpiLiFormer-4-384      & 7.04        & 4         & \textbf{81.63}        \\ \midrule
\multirow{5}{*}{CIFAR10-DVS}  & Spikformer~\cite{zhou2022spikformer} & Spikformer-2-256    & 2.57     & 16        & 80.9          \\ 
                              & Spikingformer~\cite{zhou2023spikingformer} & Spikingformer-2-256 & 2.57     & 16        & 81.3          \\ 
                              & S-Transformer~\cite{yao2024spikedriven} & S-Transformer-2-256 & 2.57     & 16        & 80.0          \\ 
                              & QKFormer~\cite{zhou2024qkformer}   & HST-2-256           & 1.50     & 16        & 84.0          \\  
                              & \textbf{SpiLiFormer (Ours)} & SpiLiFormer-2-256      & 1.57        & 16        & \textbf{86.7}          \\ \midrule
\multirow{5}{*}{N-Caltech101} & Spikformer~\cite{zhou2022spikformer} & Spikformer-2-256    & 2.57     & 16        & 83.6          \\ 
                              & Spikingformer~\cite{zhou2023spikingformer} & Spikingformer-2-256 & 2.57     & 16        & 85.91         \\ 
                              & S-Transformer~\cite{yao2024spikedriven} & S-Transformer-2-256 & 2.57     & 16        & 86.3          \\ 
                              & QKFormer~\cite{zhou2024qkformer}    & HST-2-256           & 1.50     & 16        & 87.24         \\  
                              & \textbf{SpiLiFormer (Ours)} & SpiLiFormer-2-256      & 1.57        & 16        & \textbf{89.18}         \\
\bottomrule
\end{tabular}
}
\caption{Performance comparison on CIFAR-10, CIFAR-100, CIFAR10-DVS, and N-Caltech101 datasets.}
\label{tab:small_datasets_result}
\end{table*}

\subsection{Results on ImageNet-1K}
\label{sec:imagenet_1K}

In this experiment, model training is divided into two phases. In the first phase, we use the AdamW optimizer with a base learning rate of $6 \times 10^{-4}$. The actual learning rate is calculated by multiplying the base rate by the batch size divided by 256. Additionally, we apply a layer-wise learning rate decay strategy with a decay factor of 1.0 and weight decay of 0.05 to train SpiLiFormer for 200 epochs. In the second phase, we decrease the base learning rate to $2 \times 10^{-6}$ and continue training for an additional 20 epochs using an epoch-wise learning rate decay strategy. Throughout this entire experiment, we incorporate several data augmentation techniques in alignment with the DeiT approach \cite{touvron2021training}. These techniques include RandAugment \cite{cubuk2020randaugment}, random erasing \cite{zhong2020random}, and stochastic depth \cite{huang2016deep}. The architecture of SpiLiFormer is composed of 1, 2, and 7 blocks across its three stages, respectively.

\cref{tab:imagenet_result} demonstrates the superior performance of our proposed SpiLiFormer, significantly outperforming current SNNs models. Specifically, SpiLiFormer (69.10M) achieves 86.66\% top-1 accuracy and 98.116\% top-5 accuracy on ImageNet-1K. To our best knowledge, SpiLiFormer is currently the SOTA model for Transformer-based SNNs with direct training.

We compare SpiLiFormer against several baseline spiking models under the same image resolution of 224, a time step of 4, and a comparable number of model parameters. SpiLiFormer outperforms Spikformer (66.34M), Spikingformer (66.34M), and QKFormer (64.96M) by 11.01\%, 9.97\%, and 1.6\%, respectively.
Moreover, at a higher input resolution of 288, SpiLiFormer outperforms S-Transformer and QKFormer by 9.55\% and 1.37\%, respectively. Similarly, at a resolution of 384, it surpasses QKFormer by 1.01\%.  
Additionally, we compare SpiLiFormer (69.10M, 4 time steps, 384 resolution) with E-SpikeFormer (173.0M, 8 time steps, 384 resolution), the current SOTA Transformer-based SNNs model. It achieves a 0.46\% improvement with only 39\% of parameters and half the time steps.

Finally, compared to the mainstream ANN-based Swin Transformer-B (384 resolution), SpiLiFormer achieves a 2.16\% performance improvement while reducing energy consumption by 40\% and decreasing the number of parameters by 18.67M.

\subsection{Results on CIFAR Datasets}
\label{sec:cifar_experiment}
In this experiment, we use the AdamW optimizer with a learning rate of $1 \times 10^{-3}$ and apply a cosine learning rate decay strategy. The weight decay is $6 \times 10^{-2}$, the number of time steps is 4, and SpiLiFormer is trained for 400 epochs with a batch size of 64. The network architecture consists of 1, 1, and 2 blocks across its three stages, respectively.

The performance of SpiLiFormer on CIFAR datasets is shown in the upper part of \cref{tab:small_datasets_result}. For CIFAR-10, our proposed model achieves 96.63\% accuracy with 7.04M parameters. Compared to QKFormer, the suboptimal SOTA model, SpiLiFormer improves performance by 0.45\% while increasing the parameters by only 0.04\%. Similarly, for CIFAR-100, our proposed model achieves 81.63\% accuracy with 7.04M parameters. Specifically, it outperforms QKFormer by 0.48\% while requiring just 0.04\% more parameters.

\subsection{Results on Neuromorphic Datasets}
\label{sec:neuromorphic_experiment}
In this experiment, we use a smaller SpiLiFormer architecture, structured with 0, 1, and 1 blocks across its three stages, respectively. Similarly, we adopt the AdamW optimizer with an initial learning rate of $5 \times 10^{-3}$, apply a learning rate decay strategy, and set the weight decay to 0.06. The number of time steps is set to 16, and SpiLiFormer is trained for 120 epochs.

The experimental results of the model on the neuromorphic datasets are presented in the lower half of \cref{tab:small_datasets_result}. Compared to the suboptimal model, QKFormer, SpiLiFormer significantly improves performance by 2.7\% on the CIFAR10-DVS dataset and 1.94\% on the N-Caltech101 dataset, with only a 0.13\% increase in parameters.  
Additionally, we compare the proposed model with other baseline models (\ie, Spikformer, Spikingformer, and S-Transformer). It achieves performance gains of up to 5.8\% on CIFAR10-DVS and 5.55\% on N-Caltech101 while reducing the parameter count by 0.87M.

\begin{figure}[t]
  \centering
  \includegraphics[height=10.5cm]{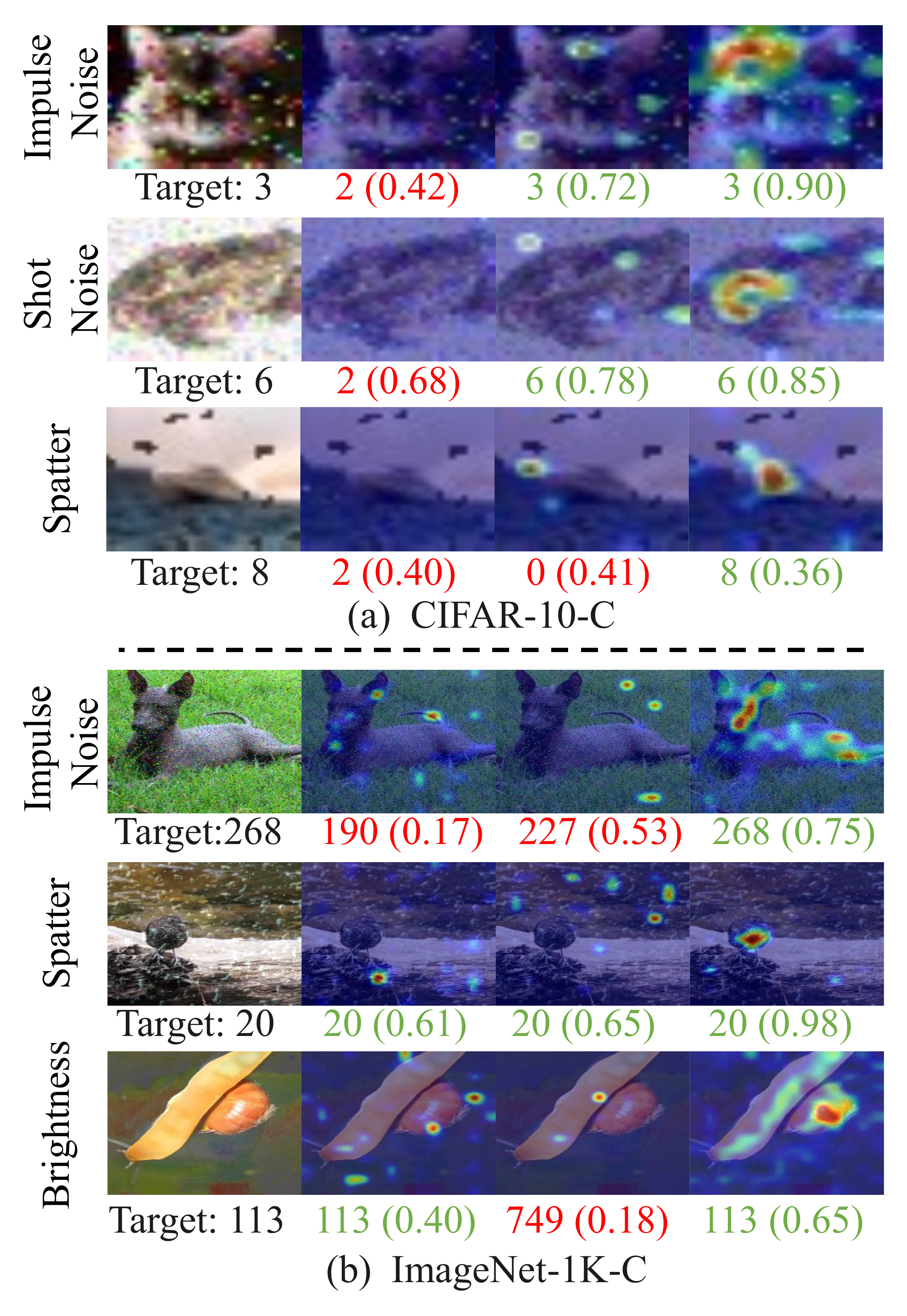}
  \caption{Visualization of corrupted versions of datasets. From left to right, we display the original image, followed by the attention heatmaps obtained from Spike-driven Transformer, QKFormer, and SpiLiFormer (ours), respectively. The information below each image includes the true label, the predicted label for each model, and the confidence score for the predicted category. Red indicates incorrect classification, while green represents correct classification. For more visualizations, refer to Fig.~6 and Fig.~7 in the appendix.}
  \label{fig:visualization_pic_noise}
\end{figure}

\subsection{Visualization and Model Robustness Analysis}

We visualize the attention heatmap of SpiLiFormer to assess whether it effectively alleviates the attention distraction issue. Specifically, we examine the attention heatmap of the last FB-LiDiff block in SpiLiFormer Stage 3 under two conditions: standard dataset evaluation and noisy dataset testing. 

In the standard dataset evaluation, the two sets of images in \cref{architecture}(a) show examples that are misclassified by the baseline model but correctly classified by SpiLiFormer on the ImageNet-1K test set. We observe that our proposed model effectively eliminates the attention distraction phenomenon, thereby improving classification accuracy. More comparison images are provided in Fig.~4 within the appendix. 

In noisy dataset testing, we evaluate model attention heatmaps using two noisy datasets: CIFAR10-C and ImageNet-1K-C, as shown in \cref{fig:visualization_pic_noise}. Compared to other baseline models, SpiLiFormer correctly classifies objects or significantly increases confidence in the correct category by focusing on object-relevant information while remaining unaffected by background noise.

\begin{figure}[t]
  \centering
  \includegraphics[width=0.47\textwidth]{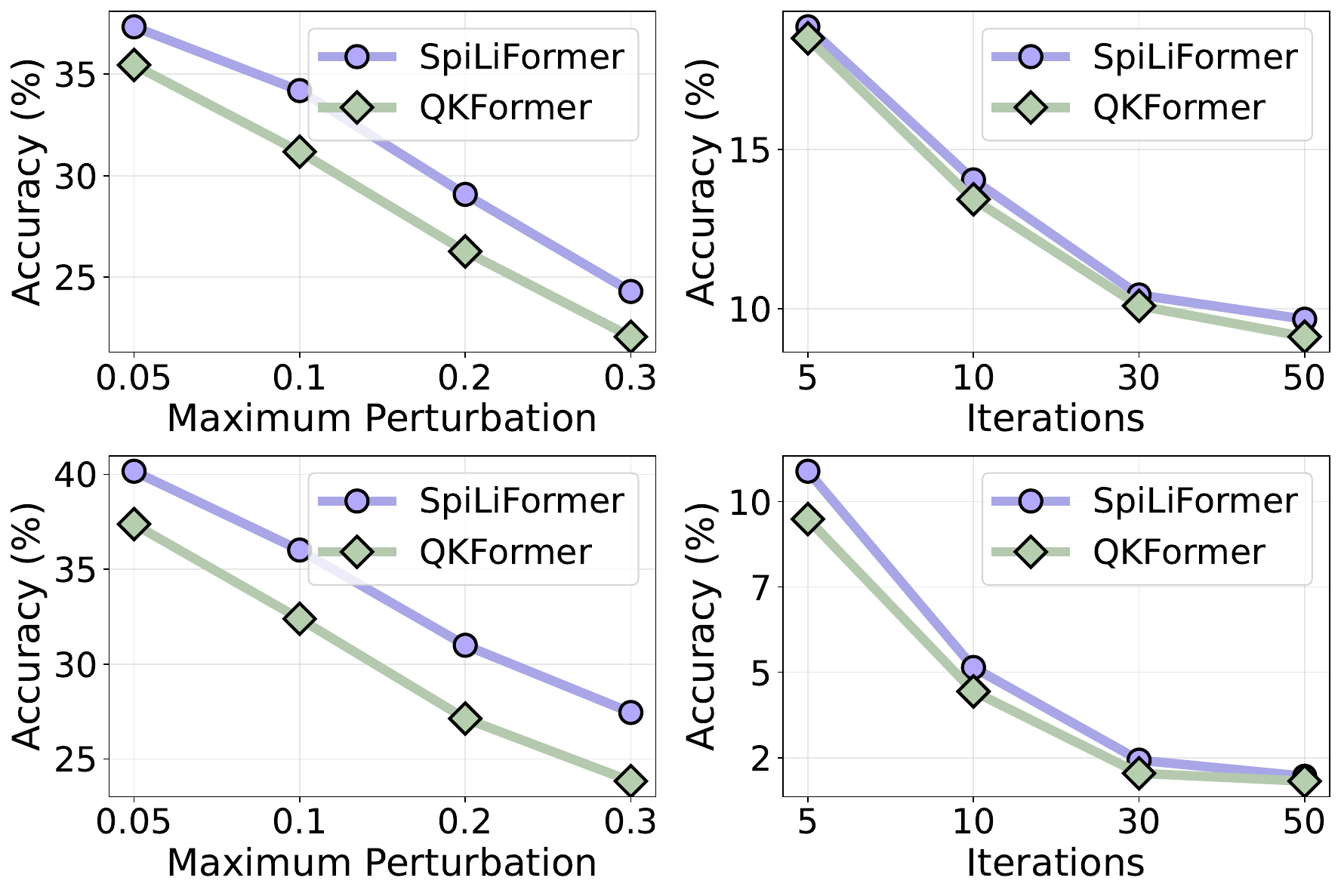}
  \caption{Adversarial robustness comparison between QKFormer and our proposed model on CIFAR-100 (first row) and ImageNet-1K (second row). For PGD, the attack strength is set to 8/255, with a step size of 2/255 per iteration. The comparison results on other datasets are provided in Tab. 6 of within the appendix.}
  \label{fig:adversarial_test}

\end{figure}

In addition, we employ white-box adversarial attacks, including FGSM and PGD, to assess the model's robustness across the four datasets mentioned above. As shown in \cref{fig:adversarial_test}, we find that SpiLiFormer not only enhances model performance but also improves robustness against adversarial and common noise attacks, outperforming QKFormer.

\subsection{Ablation Study}

We conduct an ablation study on CIFAR-10 (static) and CIFAR10-DVS (neuromorphic) to assess the impact of FF-LiDiff and FB-LiDiff attention on model performance and inference latency. All ablation experiments follow the training details in \cref{sec:cifar_experiment} and \cref{sec:neuromorphic_experiment}, unless otherwise specified.

Intuitively, we set up two experimental scenarios: 1) For the case without FF-LiDiff attention (w/o FF-LiDiff attention), we use $Attn = \mathcal{SN}(\sum_j^{D}Q^{i,j})$ to replace the steps in \cref{eq:FF_LiDiff_attn_3} and \cref{eq:FF_LiDiff_attn_4}; 2) For the case without FB-LiDiff attention (w/o FB-LiDiff attention), we perform a single forward pass with $Attn=Q\odot(K^\top \odot V)$ to calculate the attention, replacing the steps outlined in \cref{eq:FB_LiDiff_attn_4} to \cref{eq:FB_LiDiff_attn_11}. 

As shown in \cref{tab:ablation_study}, we observe that the model exhibits varying degrees of performance degradation on both datasets under the two aforementioned scenarios. Specifically, performance exhibits a decrease of 0.31\% and 0.62\% on the CIFAR-10 dataset, and 1.6\% and 3.0\% on the CIFAR10-DVS dataset. These results highlight the effectiveness of FB-LiDiff in mitigating attention distraction and improving model performance. However, its additional forward pass introduces higher inference latency, particularly on static datasets, with a 21–23\% increase as shown in Tab.~5 in the appendix.

\begin{table}[h]
\centering
\resizebox{0.45\textwidth}{!}{%
\begin{tabular}{ccc}
\toprule
Datasets & Methods & Top-1 Acc(\%) \\ \midrule
\multirow{3}{*}{CIFAR-10} & w/o FF-LiDiff Attention & 96.32 \\
 & w/o FB-LiDiff Attention & 96.01 \\
 & \textbf{SpiLiFormer(ours)} & \textbf{96.63} \\ \midrule
\multirow{3}{*}{CIFAR10-DVS} & w/o FF-LiDiff Attention & 85.1 \\
 & w/o FB-LiDiff Attention & 83.7 \\
 & \textbf{SpiLiFormer(ours)} & \textbf{86.7} \\ \bottomrule
\end{tabular}
}
\caption{Ablation study on two attention modules}
\label{tab:ablation_study}
\end{table}

\section{Conclusion}

In this paper, we identify attention distraction in mainstream Transformer-based SNNs, a critical issue that limits performance. Inspired by the brain’s lateral inhibition mechanism, we propose SpiLiFormer, incorporating FF-LiDiff and FB-LiDiff attention to simulate the inhibitory-excitatory interaction process, guiding the model to focus on the object rather than the irrelevant background. Experimental results show that SpiLiFormer outperforms current Transformer-based SNNs, achieving SOTA performance on five image classification datasets. Additionally, adversarial testing and attention visualization analysis demonstrate its robustness and effectiveness in alleviating attention distraction.
\section{Acknowledgments}
This research was supported in part by Zhejiang Provincial Natural Science Foundation of China under the Grant No.~LD25F020006.

{
    \small
    \bibliographystyle{ieeenat_fullname}
    \bibliography{main}
}

\clearpage

\appendix

\section{Appendix}

\subsection{Adversarial Test}
\label{app:a1}
Our adversarial experiments use two well-established techniques: Fast Gradient Sign Method (FGSM) and Projected Gradient Descent (PGD).

\textbf{FGSM} \cite{goodfellow2014explaining} is a single-step adversarial attack algorithm designed to generate adversarial examples efficiently. It computes the gradient of the loss function with respect to the input data and adds a small perturbation in the direction of the gradient's sign. The adversarial example $x_{\text{adv}}$ is generated as:
\begin{equation}
x_{\text{adv}} = x + \epsilon \cdot \text{sign}(\nabla_x J(x, y)) \label{eq:FGSM_appendix}
\end{equation}
where $x$ is the original input, $\epsilon$ controls the perturbation magnitude, and $J(x, y)$ is the loss function.

\textbf{PGD} \cite{mkadry2017towards} is an iterative variant of FGSM that generates  adversarial examples by repeatedly applying gradient updates. Starting from an initial perturbed input, PGD iteratively refines the perturbation while projecting the result back into a $L_{\infty}$-norm ball of radius $\epsilon$. The update rule at each iteration $t$ is:
\begin{equation}
x_{\text{adv}}^t = \text{Clip}_{x, \epsilon}\left(x_{\text{adv}}^{t-1} + \gamma \cdot \text{sign}(\nabla_x J(x_{\text{adv}}^{t-1}, y)\right) \label{eq:PGD_appendix}
\end{equation}
where $\gamma$ is the step size, and $\text{Clip}_{x, \epsilon}(\cdot)$ ensures the perturbation remains within the allowed bounds.

\subsection{Datasets}
\label{app:a3}
Our experimental evaluation includes five standard datasets, consisting of three static ones (ImageNet-1K, CIFAR-10, and CIFAR-100) and two event-based neuromorphic ones (CIFAR10-DVS and N-Caltech101).

\textbf{ImageNet-1K}: ImageNet-1K \cite{deng2009imagenet}, formally known as the ImageNet Large Scale Visual Recognition Challenge (ILSVRC) dataset, is one of the most influential benchmarks in computer vision research. It comprises over 1.28 million training images across 1,000 classes, along with 50,000 validation images and 100,000 test images.

\textbf{CIFAR-10}: CIFAR-10 \cite{krizhevsky2009learning} is a fundamental benchmark dataset in computer vision research, comprising 60,000 color images of size $32 \times 32$ pixels, distributed across 10 mutually exclusive classes.

\textbf{CIFAR-100}: CIFAR-100 \cite{krizhevsky2009learning} builds upon the design principles of CIFAR-10 while introducing a more challenging classification task. It consists of 60,000 color images of size $32 \times 32$ pixels, categorized into 100 finer-grained classes.

\textbf{CIFAR10-DVS}: CIFAR10-DVS \cite{li2017cifar10} represents a neuromorphic adaptation of the original CIFAR-10 dataset, specifically designed for the evaluation of SNNs in event-based vision tasks. There are 10,000 samples, whose spatial size is $128 \times 128$. 

\textbf{N-Caltech101}: N-Caltech101 \cite{orchard2015converting} is a neuromorphic adaptation of Caltech101, containing 101 classes and 8,709 samples with a spatial resolution of $180 \times 240$ pixels. 

\subsection{Energy Consumption Calculation of SNNs and ANNs}
\label{app:a3}

The uniformity of convolution enables the subsequent batch normalization (BN) and linear scaling transformations to be seamlessly integrated into the convolutional layer as an added bias during deployment \cite{hu2021spiking, deng2022temporal}. Consequently, when estimating theoretical energy consumption, the impact of BN layers can be disregarded. Before computing the theoretical energy consumption for SpiLiFormer, we first determine the number of Synaptic Operations (SOPs) of spikes.

\begin{equation}
  \text{SOP}^{i} = f_r \times T \times \text{FLOPs}^{i},  
  \label{eq:sop_calculate_appendix}
\end{equation}
where $i$ denotes the $i$-th layer module in SpiLiFormer, $f_r$ represents the firing rate of spike trains at the input of the layer module, and $T$ refers to the simulation time step. $\text{FLOPs}^{i}$ represents the number of floating-point operations in the $i$-th layer module, measured in terms of multiply-and-accumulate (MAC) operations. $\text{SOP}^{i}$ refers to the count of spike-based accumulate (AC) operations. We assume that MAC and AC operations are executed on 45nm hardware \cite{horowitz20141}, where $E_{\text{MAC}}=4.6pJ$ and $E_{\text{AC}}=0.9pJ$ according to previous studies \cite{yao2024spikedriven, zhou2022spikformer, kundu2021hire, horowitz20141}. The theoretical energy consumption of SpiLiFormer is computed as follows:

\begin{multline}
  E_{\text{SpiLiFormer}} = E_{\text{AC}} \times \Bigg( \sum_{i=2}^{M} \text{SOP}_{\text{Conv}}^{i} 
  + \sum_{j=1}^{N} \text{SOP}_{\text{FF-LiDiff Attn}}^{j} \\
  + \sum_{p=1}^{R} \text{SOP}_{\text{FB-LiDiff Attn}}^{p} \Bigg) 
  + E_{\text{MAC}} \times \text{FLOPs}_{\text{Conv}}^{1},  
  \label{eq:SpiLiFormer_energy_consumption_appendix}
\end{multline}
where $\text{FLOPs}_{\text{Conv}}^{1}$ represents the floating-point operations in the first convolutional layer, which processes the input image in RGB format. Subsequently, the $\text{SOPs}$ from $M$ convolutional layers, $N$ layers of FF-LiDiff attention, and $R$ layers of FB-LiDiff attention are summed and multiplied by $E_{\text{AC}}$. For ANNs, the theoretical energy consumption is determined as follows:

\begin{equation}
  E_{\text{ANN}} = E_{\text{MAC}} \times \text{FLOPs}.   
  \label{eq:anns_energy_calculate_appendix}
\end{equation}

\subsection{Selection of the Optimal \texorpdfstring{$\alpha$}{alpha} Hyperparameter}
\label{app:a4}

We perform an ablation study on both static and dynamic datasets to select the optimal value of the hyperparameter $\alpha$, as shown in Tab.~\ref{tab:hyparameter_analysis}. The results show that the model achieves peak accuracy when $\alpha = 0.5$, while other values lead to varying degrees of performance degradation. As a result, we set $\alpha$ to 0.5 by default in all subsequent experiments.

\begin{table}[h]

\centering
\resizebox{0.45\textwidth}{!}{%
\begin{tabular}{cccccc}
\toprule
\multirow{2}{*}{Datasets} & \multicolumn{5}{c}{$\alpha$  Value}      \\
& 0.3   & 0.4   & 0.5   & 0.6  & 0.7  \\
\midrule
CIFAR-10                 
& 96.35 & 96.40 & 96.63 & 96.36 & 96.16 \\
CIFAR10-DVS              
& 86.3  & 86.4  & 86.7  & 86.1 & 85.6 \\ \bottomrule
\end{tabular}
}
\caption{Ablation Study of the $\alpha$ hyperparameter}
\label{tab:hyparameter_analysis}
\end{table}

\subsection{Evaluation of Inference Latency}
\label{app:a5}

\begin{table}[h]
\centering
\resizebox{0.45\textwidth}{!}{%
\begin{tabular}{ccc}
\toprule
\multirow{2}{*}{Datasets} & \multicolumn{2}{c}{Inference Time per Sample (ms)} \\
& w/o FB-LiDiff     & w/ FB-LiDiff         \\
\midrule
CIFAR-10   & 0.7657   & 0.9322 (+21.7\%)    \\
CIFAR-100  & 0.6965   & 0.8581 (+23.2\%)     \\
CIFAR10-DVS   & 9.9433    & 10.0657 (+1.2\%)  \\
N-Caltech101   & 18.4273   & 19.8113 (+7.5\%)  \\
ImageNet-1K     & 58.9901    & 66.3027 (+12.4\%) \\     
\bottomrule
\end{tabular}
}
\caption{Inference time per sample (ms) across datasets.}
\label{tab:inference_time}
\end{table}

We conduct a comprehensive evaluation of the inference latency introduced by FB-LiDiff due to its additional forward pass. As shown in Tab.~\ref{tab:inference_time}, FB-LiDiff increases inference time by 1.2\% to 23.2\%, with over 20\% overhead observed on static CIFAR datasets.

\subsection{Supplementary Tables and Figures}
\label{app:a5}

\begin{table*}[t!]
\renewcommand{\arraystretch}{0.5}  
\centering
\resizebox{\textwidth}{!}{%
\begin{tabular}{cccccccccccc}
\toprule
\multirow{3}{*}{Dataset} & \multirow{3}{*}{Methods} & \multirow{3}{*}{Time Step} & \multirow{3}{*}{Clean} & \multicolumn{4}{c}{FGSM} & \multicolumn{4}{c}{PGD} \\
 &  &  &  & \multicolumn{4}{c}{Maximum Perturbation} & \multicolumn{4}{c}{Iterations} \\ \cmidrule{5-12}
 &  &  &  & 0.05 & 0.1 & 0.2 & 0.3 & 5 & 10 & 30 & 50 \\ \midrule
\multirow{3}{*}{CIFAR-10} & QKFormer & 4 & 96.18 & 68.33 & 65.67 & 59.51 & 53.33 & 33.94 & 25.56 & 17.96 & 16.8 \\
 & \multirow{2}{*}{SpiLiFormer(Ours)} & \multirow{2}{*}{4} & \textbf{96.63} & \textbf{68.9} & \textbf{66.05} & \textbf{59.98} & \textbf{54.46} & \textbf{35.43} & \textbf{25.7} & \textbf{18.61} & \textbf{17.13} \\
 &  &  & \textbf{(+0.45)} & \textbf{(+0.57)} & \textbf{(+0.38)} & \textbf{(+0.47)} & \textbf{(+1.13)} & \textbf{(+1.49)} & \textbf{(+0.14)} & \textbf{(+0.65)} & \textbf{(+0.33)} \\ \midrule
\multirow{3}{*}{CIFAR-100} & QKFormer & 4 & 81.15 & 35.45 & 31.18 & 26.27 & 22.07 & 18.48 & 13.43 & 10.09 & 9.12 \\
 & \multirow{2}{*}{SpiLiFormer(Ours)} & \multirow{2}{*}{4} & \textbf{81.63} & \textbf{37.33} & \textbf{34.19} & \textbf{29.08} & \textbf{24.3} & \textbf{18.84} & \textbf{14.04} & \textbf{10.43} & \textbf{9.67} \\
 &  &  & \textbf{(+0.48)} & \textbf{(+1.88)} & \textbf{(+3.01)} & \textbf{(+2.81)} & \textbf{(+2.23)} & \textbf{(+0.36)} & \textbf{(+0.61)} & \textbf{(+0.34)} & \textbf{(+0.55)} \\ \midrule
\multirow{3}{*}{CIFAR10-DVS} & QKFormer & 16 & 84.00 & 29.30 & 17.70 & 10.50 & 10.00 & 2.00 & 1.20 & 0.40 & \textbf{0.50} \\
 & \multirow{2}{*}{SpiLiFormer(Ours)}  & \multirow{2}{*}{16} & \textbf{86.70} & \textbf{34.50} & \textbf{23.70} & \textbf{15.60} & \textbf{11.70} & \textbf{3.50} & \textbf{1.40} & 0.40 & 0.40 \\
 &  &  & \textbf{(+2.70)} & \textbf{(
 +5.20)} & \textbf{(+6.00)} & \textbf{(+5.10)} & \textbf{(+1.70)} & \textbf{(+1.50)} & \textbf{(+0.20)} & 0.00 & -0.10 \\ \midrule
\multirow{3}{*}{ImageNet-1K} & QKFormer & 1 & 80.10 & 37.38 & 32.39 & 27.13 & 23.83 & 9.49 & 4.44 & 2.04 & 1.81 \\
 & \multirow{2}{*}{SpiLiFormer(Ours)} & \multirow{2}{*}{1} & \textbf{81.54} & \textbf{40.17} & \textbf{36.01} & \textbf{30.99} & \textbf{27.45} & \textbf{10.89} & \textbf{5.14} & \textbf{2.43} & \textbf{1.95} \\
 &  &  & \textbf{+(1.44)} & \textbf{(+2.79)} & \textbf{(+3.62)} & \textbf{(+3.86)} & \textbf{(+3.62)} & \textbf{(+1.40)} & \textbf{(+0.70)} & \textbf{(+0.39)} & \textbf{(+0.14)} \\ \bottomrule
\end{tabular}
}
\caption{Adversarial robustness comparison between QKFormer and our model across four datasets, including CIFAR-10, CIFAR-100, CIFAR10-DVS, and ImagetNet-1K. For PGD, the attack strength is set to 8/255, with a step size of 2/255 per iteration.}
\label{tab:adv_test}
\end{table*}

\begin{figure*}[t]
    \centering
    \includegraphics[width=0.80\textwidth]{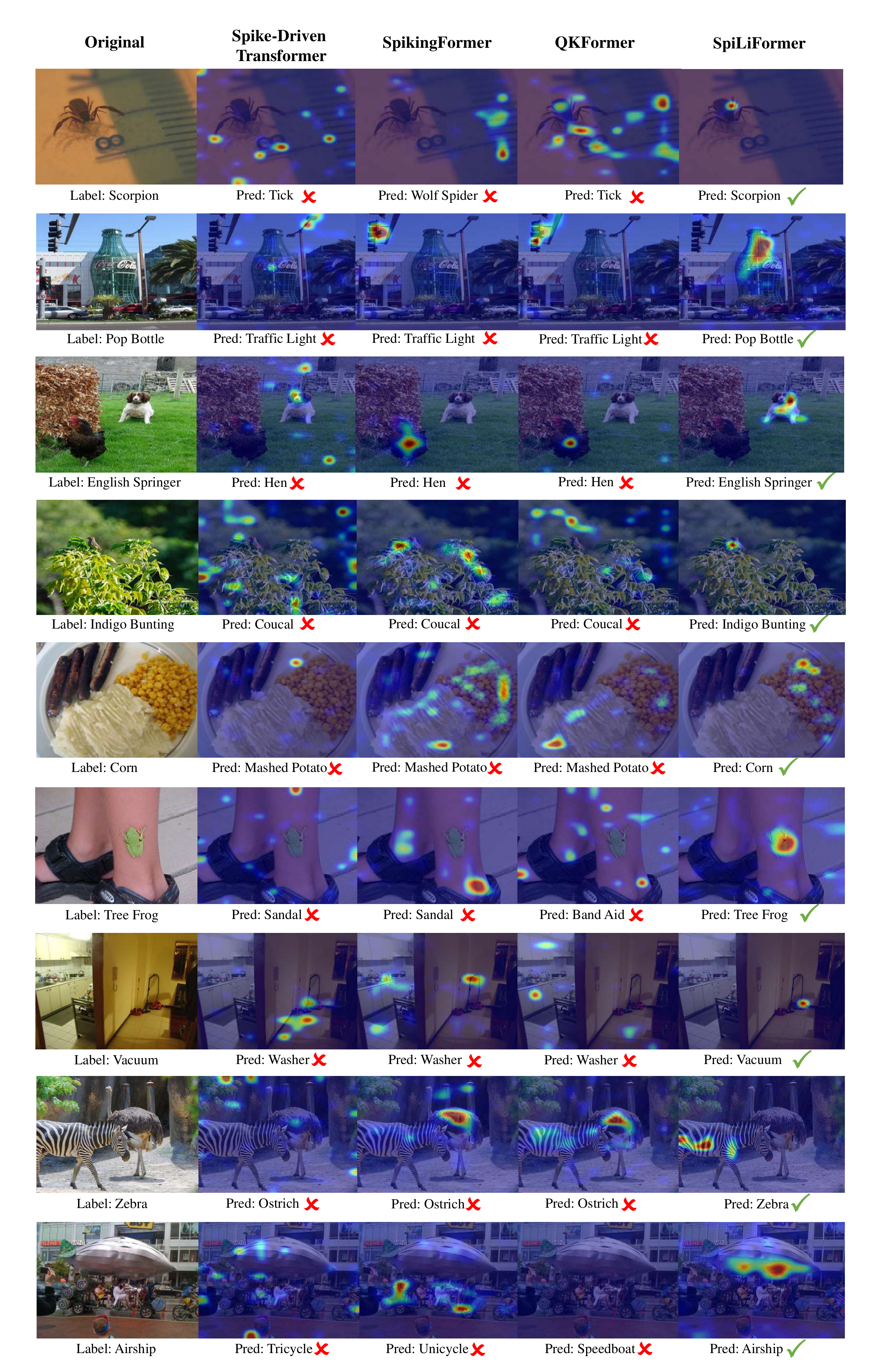} 
    \caption{Comparative visualization of attention heatmaps from ImageNet-1K, with corresponding ground truth labels and model predictions annotated below each sample.} 
    \label{fig:imagetnet_1K_others_wrong_ours_right_appendix}
\end{figure*}

\begin{figure*}[t]
    \centering
    \includegraphics[width=0.60\textwidth]{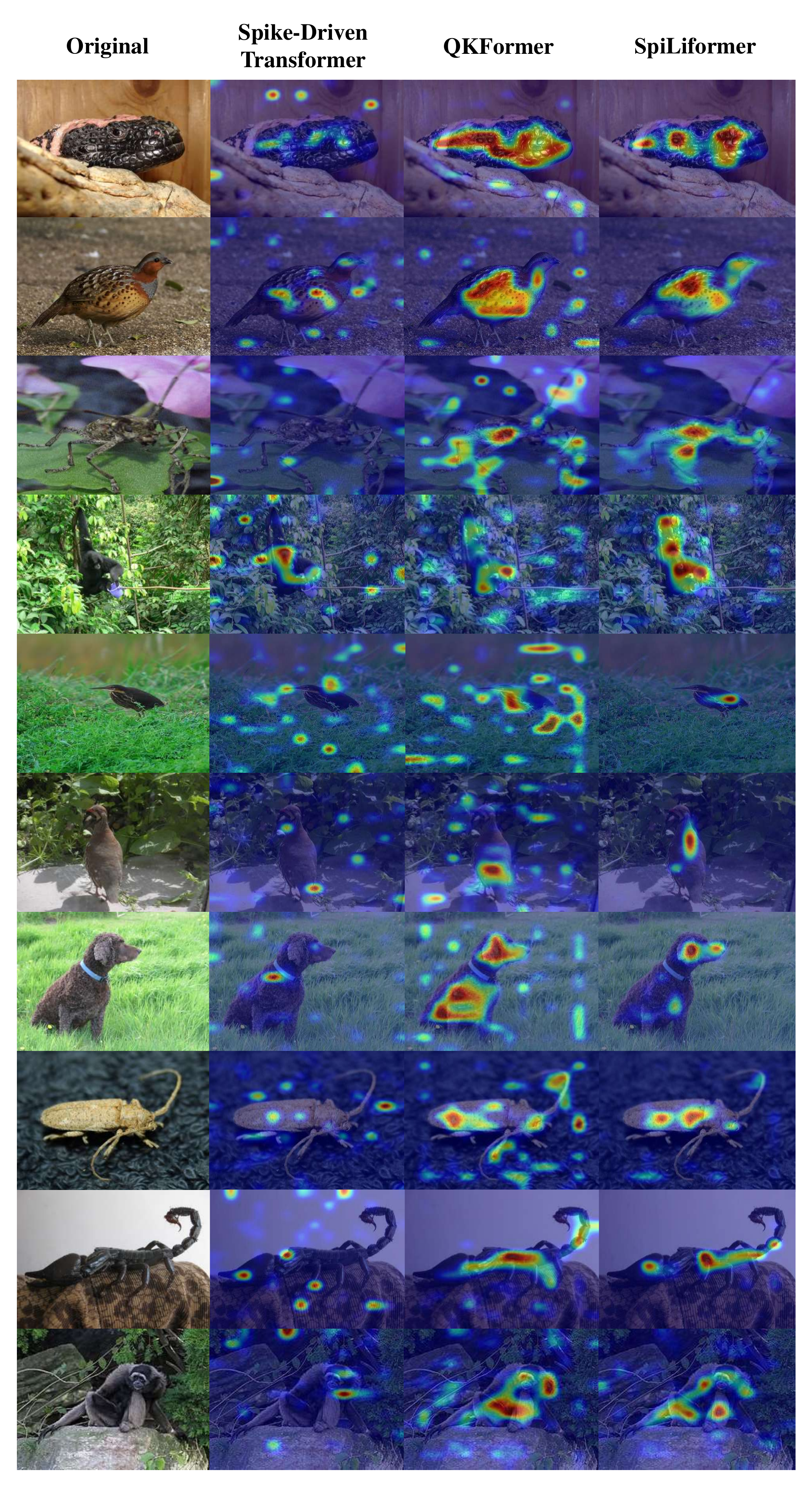} 
    \caption{Representative samples from ImageNet-1K, demonstrating  original images and their corresponding attention heatmaps across different models.} 
    \label{fig:imagenet_1K_attention_appendix}
\end{figure*}

\begin{figure*}[t]
    \centering
    \includegraphics[width=0.60\textwidth]{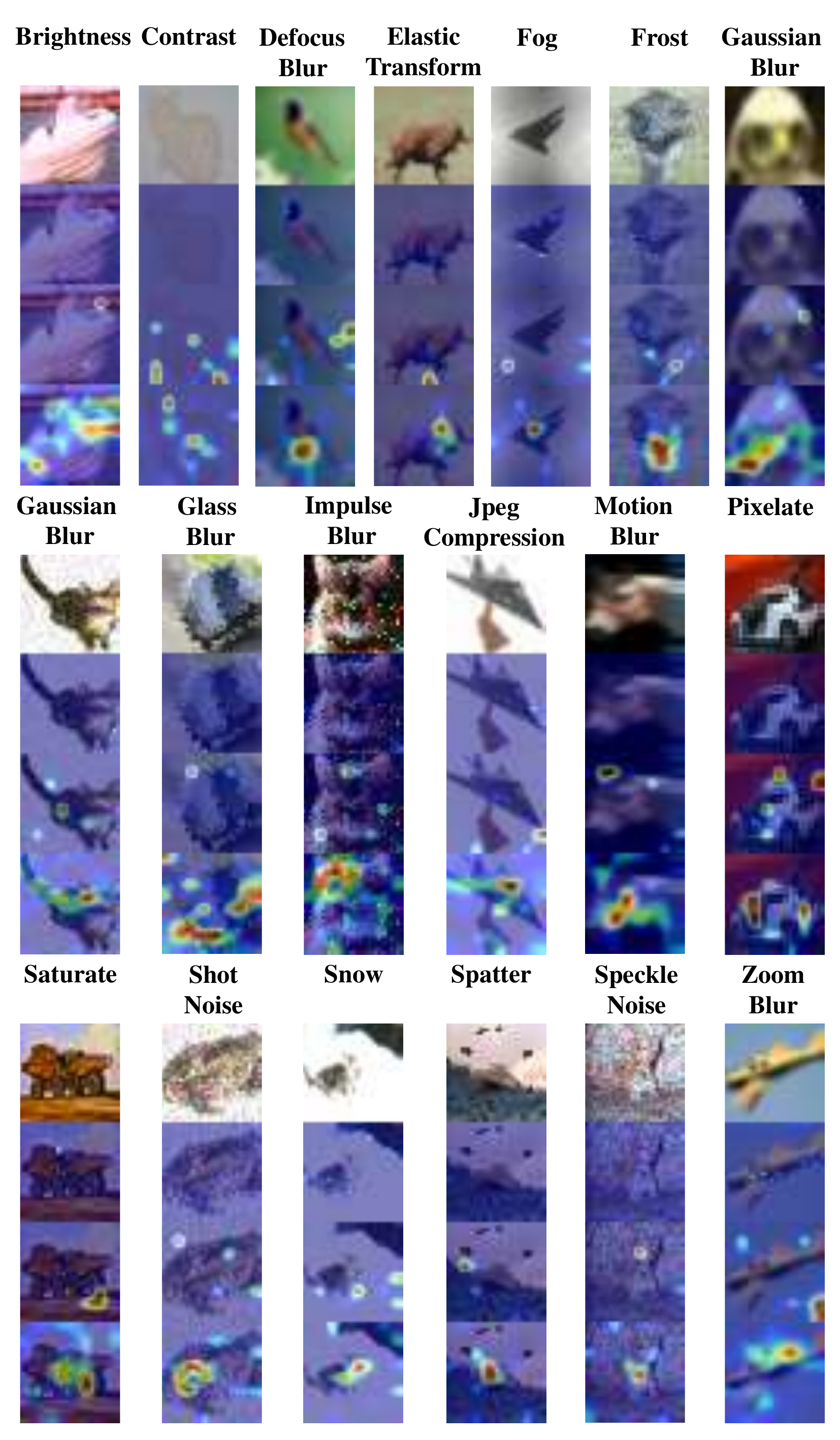} 
    \caption{Visualization of CIFAR-10C. For all 19 types of corruptions, each column displays the following cases: the first image is the original corrupted image; the second and third images show the attention heatmaps of Spike-Driven Transformer and QKFormer, respectively; the last image visualizes the attention of SpiLiFormer.} 
    \label{fig:cifar_10_C_appendix}
\end{figure*}

\begin{figure*}[t]
    \centering
    \includegraphics[width=0.60\textwidth]{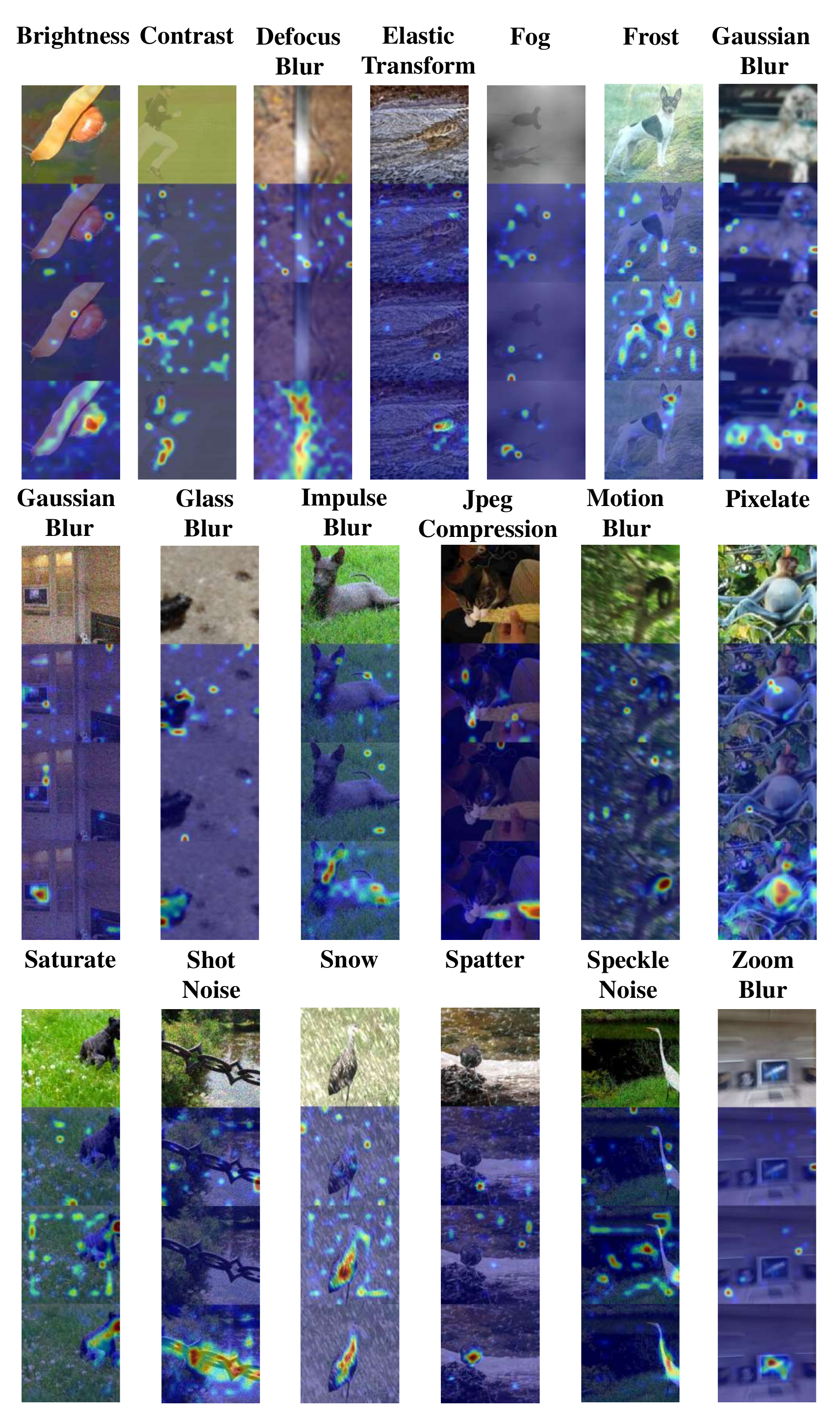}
    \caption{Visualization of ImageNet-1K-C for all types of corruptions. The layout and image order follow the same structure as illustrated in \cref{fig:cifar_10_C_appendix}.} 
    \label{fig:imagnet_1K_C_appendix}
\end{figure*}


\end{document}